\documentclass[runningheads]{llncs}
\usepackage{graphicx}
\usepackage{amsmath,amssymb} 
\usepackage{color}
\usepackage{multirow} 
\usepackage{xspace}
\usepackage{capt-of}
\usepackage{enumitem}
\usepackage[width=122mm,left=12mm,paperwidth=146mm,height=193mm,top=12mm,paperheight=217mm]{geometry}
\usepackage[hyphens]{url}
\usepackage[table]{xcolor}
\usepackage{fancyvrb}

\begin{document}

\Urlmuskip=0mu plus 1mu\relax

\newcommand{\lrfont}[1]{ \parbox[t]{5cm}{ {#1}} }

\newcommand{\layerfont}[1]{ \multicolumn{1}{|l|}{ {#1}} }  
\newcommand{\expertlayerfont}[1]{ \multicolumn{1}{|l|}{ \cellcolor{yellow!30}  {#1}} }  %

\newcommand{\R}{\mathbb R}
\newcommand{\bx}{\mbox{\boldmath $x$}}
\def\nofe{{\scshape NofE}\xspace}
\def\myparagraph#1{\noindent{\bf #1~~}}
\def\LT#1{{{{\color{green} \it #1}}}}

\def\HB#1{{{{\color{blue} \it #1}}}}
\def\KA#1{{{{\color{cyan} \it #1}}}}

\def\LTR#1{}

\pagestyle{headings}
\mainmatter

\title{Network of Experts for\\Large-Scale Image Categorization} 

\titlerunning{Network of Experts for Large-Scale Image Categorization}

\authorrunning{Ahmed, Baig, Torresani}

\author{Karim Ahmed \and Mohammad Haris Baig \and Lorenzo Torresani}


\institute{Department of Computer Science,\\
	Dartmouth College\\
	\email{ \{karim, haris\}@cs.dartmouth.edu, LT@dartmouth.edu}
}


\maketitle

\begin{abstract}
We present a tree-structured network architecture for large-scale image classification. The trunk of the network contains convolutional layers optimized over all classes. At a given depth, the trunk splits into separate branches, each dedicated to discriminate a different subset of classes. Each branch acts as an expert classifying a set of categories that are difficult to tell apart, while the trunk provides common knowledge to all experts in the form of shared features. The training of our ``network of experts'' is completely end-to-end: the partition of categories into disjoint subsets is learned simultaneously with the parameters of the network trunk and the experts are trained jointly by minimizing a single learning objective over all classes. The proposed structure can be built from any existing convolutional neural network (CNN). We demonstrate its generality by adapting 4 popular CNNs for image categorization into the form of networks of experts. Our experiments on CIFAR100 and ImageNet show that in every case our method yields a substantial improvement in accuracy over the base CNN, and gives the best result achieved so far on CIFAR100. Finally, the improvement in accuracy comes at little additional cost: compared to the base network, the training time is only moderately increased and the number of parameters is comparable or in some cases even lower.  Our code is available at: {\bf \color{red} \url{http://vlg.cs.dartmouth.edu/projects/nofe/}}

\end{abstract}

\section{Introduction}
Our visual world encompasses tens of thousands of different categories. While a layperson can recognize effectively most of these visual classes~\cite{Biederman:PsychologicalRev1987}, discrimination of categories in specific domains requires expert knowledge that can be acquired only through dedicated training. Examples include learning to identify mushrooms, authenticate art, diagnosing diseases from medical images. In a sense, the visual system of a layperson is a very good generalist that can accurately discriminate coarse categories but lacks the specialist eye to differentiate fine categories that look alike. Becoming an expert in any of the aforementioned domains involves time-consuming practical training aimed at specializing our visual system to recognize the subtle features that differentiate the given classes. 

Inspired by this analogy, we propose a novel scheme that decomposes large-scale image categorization into two separate tasks: 1) the learning of a generalist optimized to discriminate coarse groupings of classes, i.e., disjoint subsets of categories which we refer to as ``specialties'' and 2) the training of experts that learn specialized features aimed at accurate recognition of classes within each specialty. Rather than relying on a hand-designed partition of the set of classes, we propose to {\em learn} the specialties for a substantial improvement in accuracy (see Fig.~\ref{fig:specialtyacc}). Our scheme simultaneously learns the specialties and the generalist that is optimized to recognize these specialties. We frame this as a joint minimization of a loss function $E(\theta^G, \ell)$ over the parameters $\theta^G$ of the generalist and a labeling function $\ell$ that maps each original category to a specialty. 
In a second training stage, for each specialty, an expert is trained to classify the categories within that specialty.

\begin{figure}[t!]
\centering
\includegraphics[height=5.5cm]{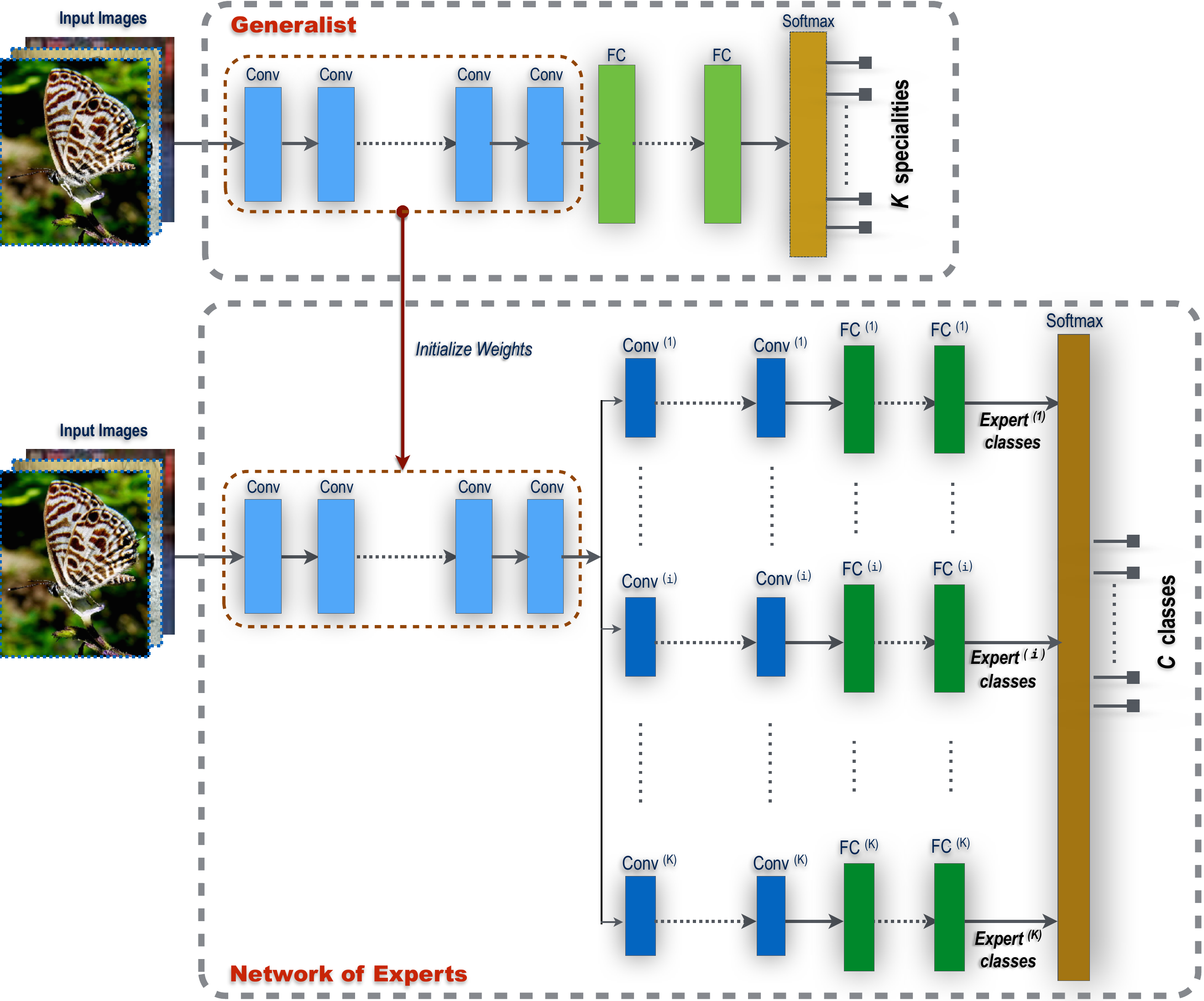}
\caption{Our Network of Experts (\nofe). {\bf Top:} Training of the generalist. The generalist is a traditional CNN but it is optimized to partition the original set of $C$ classes into $K<<C$ disjoint subsets, called specialties. Our method performs {\em joint} learning of the $K$ specialties and the generalist CNN that is optimized to recognize these specialties. {\bf Bottom:} The complete \nofe with $K$ expert branches. The convolutional layers of the generalist are used as initialization for the trunk, which ties into $K$ separate branches, each responsible to discriminate the classes within a specialty. The complete model is trained end-to-end via backpropagation with respect to the original $C$ classes.}
\label{fig:model}
\end{figure}

Although our learning scheme involves two distinct training stages -- the first aimed at learning the generalist and the specialties, the second focused on training the experts -- the final product is a unified model performing multi-class classification over the original classes, which we call ``Network of Experts'' (\nofe). The training procedure is illustrated in Fig~\ref{fig:model}. The generalist is implemented in the form of a convolutional neural network (CNN) with a final softmax layer over $K$ specialties, where $K << C$, with $C$ denoting the original number of categories (Figure~\ref{fig:model}(top)). 
After this first training stage, the fully connected layers are discarded and $K$ distinct branches are attached to the last convolutional layer of the generalist, i.e., one branch per specialty. 
Each branch is associated to a specialty and is devoted to recognize the classes within the specialty. This gives rise to the \nofe architecture, a unified tree-structured network (Figure~\ref{fig:model}(bottom)). 
Finally, all layers of the resulting model are fine-tuned with respect to the original $C$ categories by means of a global softmax layer that calibrates the outputs of the individual experts over the $C$ categories.

Thus, the learning of our generalist serves two fundamental purposes:

\noindent 1) First, using a {\em divide and conquer} strategy it decomposes the original multi-class classification problem over $C$ labels into $K$ subproblems, one for each specialty. The specialties are defined so that the act of classifying an image into its correct specialty is as accurate as possible. At the same time this implies that confusable classes are pushed into the same specialty, thus handing off the most challenging class-discrimination cases to the individual experts. However, because each expert is responsible for classification only over a subset of classes that are highly similar to each other, it can learn highly specialized and effective features for the subproblem, analogously to a human expert identifying mushrooms by leveraging features that are highly domain-specific (cap shape, stem type, spore color, flesh texture, etc.). 

\noindent 2) Second, the convolutional layers learned by the generalist provide an initial knowledge-base for all experts in the form of shared features. In our experiments we demonstrate that fine-tuning the trunk from this initial configuration results in a significant improvement over learning the network of experts from scratch or even learning from a set of convolutional layers optimized over the entire set of $C$ labels. Thus, the subproblem decomposition does not merely simplify the original hard classification problem but it also produces a set of pretrained features that lead to better finetuning results. 

We note that we test our approach on image categorization problems involving a large number of classes, such as ImageNet classification, where the classifier must have the ability to recognize coarse categories (``vehicle'') but must also distinguish highly confusable specialty classes (e.g., ``English pointer'' from ``Irish setter''). These scenarios match well the structure of our model, which combines a generalist with a collection of experts. We do not assess our approach on a fine-grained categorization benchmark as this typically involves classification focused only on one domain (say, bird species) and thus does not require the generalist and multiple specialists learned by our model.


\section{Related Work}

Our work falls in the category of CNN models for image classification. This genre has witnessed dramatic growth since the introduction of the ``AlexNet'' network~\cite{AlexNet}. In the last few years further recognition improvements have been achieved thanks to advances in CNN components~\cite{GlorotEtAl:AISTATS2011,MaasEtAl:ICML2013,HeEtAl:ICCV2015,HeEtAl:ECCV2014} and training strategies~\cite{LeeEtAL:ICML2009,ErhanEtAl:JMLR2010,LeeEtAl:AISTATS,IoffeSzegedy:arXiv2015,HintonEtAl:arXivDropout2012}. Our approach instead achieves gains in recognition accuracy by means of an architectural alteration that involves adapting existing CNNs into a tree-structure. Thus, our work relates to prior studies of how changes in the network structure affect performance~\cite{KontschiederEtAl:ICCV2015,HeEtAl:arXivResNet}.

Our adaptation of base CNN models into networks of experts hinges on a method that groups the original classes into specialties, representing subsets of confusable categories. Thus, our approach relates closely to methods that learn hierarchies of categories. This problem has received ample study, particularly for the purpose of speeding up multi-class classification in problems involving large number of categories~\cite{GriffinPerona:CVPR2008,MarszalekSchmid:ECCV2008,BengioEtAl:NIPS2010,GaoKoller:ICCV2011,DengEtAl:CVPR2012,LiuEtAl:CVPR2013}. 
Hierarchies of classes have also been used to infer class abstraction~\cite{JiaEtAl:NIPS2013}, to trade off concept specificity versus accuracy~\cite{DengEtAl:CVPR2012,OrdonezEtAl:ICCV2013},  to allow rare objects to borrow statistical strength from related but more frequent objects~\cite{SalakhutdinovEtAl:CVPR2011,SrivastavaSalakhutdinov:NIPS2013}, and also for unsupervised discovery of objects~\cite{SivicEtAl:CVPR2008}.

Our proposed work is most closely related to methods that learn groupings of categories in order to train expert CNNs that specialize in different visual domains. Hinton et al.~\cite{HintonEtAl:arXivDISTILL} introduced an ensemble network composed of one or more full models and many specialist models which learn to distinguish fine-grained classes that the full models confuse. Similarly, Warde-Farley et al.~\cite{FarleyEtAl:ICLR2015} augment a very large CNN trained over all categories via auxilliary hidden layer pathways that connect to specialists trained on subsets of classes. Yan et al.~\cite{HD-CNNICCV15} presented a hierarchical deep CNN (HD-CNN) that consists of a coarse component trained over all classes as well as a set of fine components trained over subsets of classes. The coarse and the fine components share low-level features and their predictions are late-fused via weighted probabilistic averaging. While our approach is similar in spirit to these three expert-systems, it differs substantially in terms of architecture and it addresses some of their shortcomings:
\begin{enumerate}
\item In~\cite{HintonEtAl:arXivDISTILL,FarleyEtAl:ICLR2015,HD-CNNICCV15} the experts are learned only after having trained a large-capacity CNN over the original multi-class classification problem. 
The training of our approach does not require the expensive training of the base CNN model over all $C$ classes. Instead it directly learns a generalist that discriminates a much smaller number of specialties ($K << C$). The experts are then trained as categorizers within each specialty. By using this simple {\em divide and conquer} the training cost remains manageable and the overall number of parameters can be even lower than that of the base model (see Table~\ref{table:accuracy_imagenet_alexnet_val}). 
\item The architectures in~\cite{HintonEtAl:arXivDISTILL,HD-CNNICCV15} route the input image to only a subset of experts, those deemed more competent in its categorization. A routing mistake cannot be corrected. To minimize this error, redundancy between experts must be built by using overlapping specialties (i.e., specialties sharing classes) thus increasing the number of classes that each expert must recognize. Instead, in our approach the specialties are disjoint and thus more specific. Yet, our method does not suffer from routing errors as all experts are invoked in parallel for each input image.
\item Although our training procedure involves two distinct stages, the final phase performs fine-tuning of the complete network of experts using a single objective over the original $C$ categories. While fine-tuning is in principle possible for both~\cite{HintonEtAl:arXivDISTILL} and~\cite{FarleyEtAl:ICLR2015}, in practice this was not done because of the large computational cost of training and the large number of parameters.
\end{enumerate}


\section{Technical approach}

In this section we present the details of our technical approach. We begin by introducing the notation and the training setup. 

Let $\mathcal{D} = \{(x^1,y^1), \hdots, (x^N,y^N)\}$ be a training set of $N$ class-labeled images where $x^i \in \R^{r \times c \times 3}$ represents the $i$-th image (consisting of $r$ rows, $c$ columns and 3 color channels) and $y^i \in \mathcal{Y} \equiv \{1, 2, \hdots, C\}$ denotes its associated class label ($C$ denotes the total number of classes). 

Furthermore, we assume we are given a CNN architecture $b_{\theta^B}: \R^{r \times c \times 3} \longrightarrow Y$ parameterized by weights $\theta^B$ that can be optimized to categorize images into classes $Y$. We refer to this CNN model as the {\em base} architecture, since we will use this architecture to build our network of experts resulting in a classifier $e_{\theta^E}: \R^{r \times c \times 3} \longrightarrow Y$. In our empirical evaluation we will experiment with different choices of base classifiers~\cite{AlexNet,SimonyanZisserman:ICLR2015,HeEtAl:arXivResNet}. Here we abstract away the specificity of individual base classifiers by assuming that $b_{\theta^B}$ consists of a CNN with a certain number of convolutional layers followed by one or more fully connected layers and a final softmax layer that defines a posterior distribution over classes in $\mathcal{Y}$. Finally, we assume that the parameters of the base classifier can be learned by optimizing an objective function of the form:
\begin{equation}
E_b(\theta; \mathcal{D}) = R(\theta) + \frac{1}{N} \sum_{i=1}^N L(\theta; x^i, y^i)
\label{eq:baselearning}
\end{equation}
where $R$ is a regularization term aimed at preventing overfitting (e.g., weight decay) and $L$ is a loss function penalizing misclassification (e.g., the cross entropy loss). As $\mathcal{D}$ is typically large, it is common to optimize this objective using backpropagation over mini-batches, i.e., by considering at each iteration a random subset of examples $\mathcal{S} \subset \mathcal{D}$ and then minimizing $E_b(\theta; \mathcal{S})$.

In the following subsections we describe how to adapt the architecture of the base classifier $b_{\theta^B}$ and the objective function $E_b$ in order to learn a network of experts. Note that our approach does not require {\em learning} (i.e., optimizing) the parameters of the base classifier (i.e., optimizing parameters $\theta^B$). Instead it simply needs a base CNN architecture (with uninstantiated weights) and a learning objective. We decompose the training into two stages: the learning of the generalist (described in subsection~\ref{sec:generalist}) and the subsequent training of the complete network of experts (presented in subsection~\ref{sec:experts}), which uses the generalist as initialization for the trunk and the definition of the specialties. 

\subsection{Learning the Generalist}
\label{sec:generalist}

The goal of this first stage of training is to learn groupings of classes, which we call specialties. Intuitively, we want each specialty to represent a subset of classes that are highly confusable (such as different mushrooms) and that, as such, require the specialized analysis of an expert. Formally, the specialties represent a partition of the set $Y$. In other words, the specialties are $K << C$ disjoint subsets of classes whose union gives $\mathcal{Y}$ and where $K$ represents a hyperparameter defining the number of experts and thus the complexity of the system. We can cast the definition of the specialties as the problem of learning a label mapping $\ell: \mathcal{Y}  \longrightarrow \mathcal{Z}$, where $\mathcal{Z} = \{1, 2, \hdots, K\}$ is the set of specialty labels. Conceptually we want to define $\ell$ such that we can train a generalist $g_{\theta^G}: \R^{r \times c \times 3}  \longrightarrow \mathcal{Z}$ that correctly classifies image $x^i$ into its associated specialty, i.e., such that $g(x^i; \theta^G) = \ell(y^i)$. 
We formulate this task as a joint optimization over the parameters  $\theta^G$ of the generalist and the mapping $\ell$ so as to produce the best possible recognition accuracy over the specialty labels by considering the objective
\begin{equation}
E_g(\theta^G, \ell; \mathcal{D}) = R(\theta) + \frac{1}{N} \sum_{i=1}^N L(\theta; x^i, \ell(y^i)).
\end{equation}
Note that this is the same objective as in Eq.~\ref{eq:baselearning}, except that the labels of the examples are now defined in terms of the mapping $\ell$, which is itself unknown.  Thus we can now view this learning objective as a function over unknown parameters $\theta^G, \ell$. The architecture of the generalist is the same as that of the base model except for the use of a softmax over $\mathcal{Z}$ instead of $\mathcal{Y}$  and for the dimensionality of the last fully connected layer, which also needs to change in order to match the number of specialties, $K$.

We optimize this objective via a simple alternation scheme that iterates between the following two steps: 
\begin{enumerate}
\item Optimizing parameters $\theta^G$ while keeping specialty labels $\ell$ fixed.
\item Updating specialty labels $\ell$ given the current estimate of weights $\theta^G$. 
\end{enumerate}
First, we initialize the mapping $\ell$ by randomly partitioning $Y$ into $K$ subsets, each containing $C/K$ classes (in all our experiments we use values of $K$ that are factors of $C$ so that we can produce a set of $K$ perfectly-balanced specialties). Given this initial set of specialty labels, the first step of the alternation scheme is implemented by running several iterations of stochastic gradient descent.

The second step of our alternation requires optimizing with respect to $\ell$ given the current estimate of parameters $\theta^G$. For this purpose, we evaluate the generalist defined by the current parameters $\theta^G$ over a random subset $\mathcal{S} \subset \mathcal{D}$ of the training data. For this set we build the confusion matrix $M \in \R^{C \times K}$, where $M_{ij}$ is the fraction of examples of class label $i$ that are classified into specialty $j$ by the {\em current} generalist. Then, a greedy strategy would be to set $\ell(i) = \arg \max_{j \in \{1,\hdots,K\}} M_{ij}$ for each class $i \in \{1, \hdots, C\}$ so that each class is assigned to the specialty that recognizes the maximum number of images of that class. Another solution is to perform spectral clustering over the confusion matrix, as done in~\cite{HD-CNNICCV15,HintonEtAl:arXivDISTILL,bergamo:cvpr12}. However, we found that both of these solutions yield highly imbalanced specialty clusters, where a few specialties absorb nearly all  classes, making the problem of classification within these large specialties almost as hard as the original, as confirmed in our experiments. To address this problem we tried two different schemes that either constrain or softly encourage the specialties to have an equal number of classes, as discussed next.
\begin{itemize}[leftmargin=*]
\item The first scheme, which we refer to as {\tt fully-balanced} forces the specialties to have equal size. Initially the specialties are set to be empty and they are then grown by considering the classes in $\mathcal{Y}$ one at a time, in random order. For each class $i \in \mathcal{Y}$, we assign the specialty $j$ that has the highest value $M_{i,j}$ among the specialties that have not yet reached maximum size $C/K$. The randomization in the visiting order of classes guarantees that, over multiple label updates, no class is favored over the others.
\item Unlike the previous scheme which produces perfectly balanced specialties, {\tt elasso} is a method that allows us to encourage {\em softly} the constraint over the size of specialties. This may be desirable in scenarios where certain specialties should be allowed to include more classes than others. 
The procedure is adapted from the algorithm of Chang et al.~\cite{elassoclusteringpaper}. To define the specialties for this method we use a clustering indicator matrix $F \in \{0, 1\}^{C \times K}$ where each row of $F$ has one entry only set to 1, denoting the specialty assigned to the class. Let us indicate with $Ind(C,K)$ the set of all clustering indicator matrices of size $C \times K$ that satisfy this constraint. In order to create specialties that are simultaneously easy to classify and balanced in size, we compute $F$ by minimizing the objective
\begin{equation}
\min_{F \in Ind(C,K)} \lambda || F ||_e - || M \odot F ||_{1,1} 
\label{eq:elasso}
\end{equation}
where $|| F ||_e = \sqrt{\sum_{j} \left( \sum_i F_{ij} \right)^2}$ is the so-called exclusive lasso norm~\cite{elassoclusteringpaper,ZhouEtAl}, $\odot$ denotes the element-wise product between matrices, $||A||_{1,1} = \sum_i \sum_j |A_{ij}|$ is the $L_{1,1}$-norm, and $\lambda$ is a hyperparameter trading off the importance between having balanced specialties and good categorization accuracy over them. Note that the first term captures the balance degree of the specialties: for each $j$, it computes the squared-number of classes assigned to specialty $j$ and then sums these squared-numbers over all specialties. Thus, $|| F ||_e$ uses an $L_1$-norm to compute the number of classes assigned to each specialty, and then an $L_2$-norm to calculate the average size of the specialty. The $L_2$-norm strongly favors label assignments that generate specialties of roughly similar size. The second term, $|| M \odot F ||_{1,1} = \sum_{j} \sum_{i} M_{ij} F_{ij}$, calculates the {\em accuracy} of specialty classification. As we want to make specialty classification accuracy as high as possible, we subtract this term from the exclusive lasso norm to define a {\em minimization} objective. As in~\cite{elassoclusteringpaper}, we update one row of $F$ at a time. Starting from an initial $F$ corresponding to the current label mapping $\ell$, we loop over the rows of $F$ in random order and for each row we find the element being 1 that yields the minimum of Eq.~\ref{eq:elasso}. This procedure is repeated until convergence (see~\cite{elassoclusteringpaper} for a proof of guaranteed convergence).
\end{itemize}

\subsection{Training the Network of Experts}
\label{sec:experts}

Given the generalist $\theta^G$ and the class-to-specialty mapping $\ell$ produced by the first stage of training, we perform joint learning of the $K$ experts in order to obtain a global multi-class classification model over the original categories in the label set $\mathcal{Y}$. As illustrated in~Fig.\ref{fig:model}, this is achieved by defining a tree-structured network consisting of a single trunk feeding $K$ branches, one branch for each specialty. The trunk is initialized with the convolutional layers of the generalist, as they have been optimized to yield accurate specialty classification. Each branch contains one or more convolutional layers followed by a number of fully-connected layers (in our experiments we set the expert to have as many fully connected layers as the base model). However, each branch is responsible for discriminating only the classes associated to its specialty. Thus, the number of output units of the last fully-connected layer is equal to the number of classes in the specialty (this is exactly equal to $C/K$ for {\tt fully-balanced}, while it varies for individual specialties in {\tt elasso}). The final fully-connected layer of each branch is fed into a {\em global} softmax layer defined over the entire set of $C$ labels in the set $\mathcal{Y}$. This softmax layer does not contain weights. Its purpose is merely to normalize the outputs of the $K$ experts to define a proper class posterior distribution over $\mathcal{Y}$.

The parameters of the resulting architecture are optimized via backpropagation with respect to the training set $\mathcal{D}$ and labels in $\mathcal{Y}$ using the regularization term $R$ and loss function $L$ of the base model.  
This implies that for each example $x^i$ both forward and backward propagation will run in all $K$ branches, irrespective of the ground truth specialty $\ell(y^i)$ of the example. The backward pass from the ground-truth branch will aim at increasing the output value of the correct class $y^i$, while the backward pass from the other $K-1$ branches will update the weights to lower the probabilities of their classes. Because of this joint training the outputs of the experts will be automatically calibrated to define a proper probability distribution over $\mathcal{Y}$. While the weights in the branches are randomly initialized, the convolutional layers in the trunk are initially set to the parameters computed by the generalist. Thus, this final learning stage can be viewed as performing  fine-tuning of the generalist parameters for the classes in $\mathcal{Y}$ using the network of experts. 

Given the learned \nofe, inference is done via forward propagation through the trunk and all $K$ branches so as to  produce a full distribution over $\mathcal{Y}$.

\section{Experiments}

We performed experiments on two different datasets: CIFAR100~\cite{Krizhevsky:TR2009}, which is a medium size dataset, and the large-scale ImageNet benchmark~\cite{ImageNet}. 

\subsection{Model Analysis on CIFAR100}
\label{exp:cifar100}

The advantage of CIFAR100 is that its medium size allows us to carry out a comprehensive study of many different design choices and architectures, which would not be feasible to perform on the large-scale ImageNet benchmark. CIFAR100 consists of color images of size 32x32 categorized into 100 classes. The training set contains 50,000 examples (500 per class) while the test set consists of 10,000 images (100 per class). 

Our first set of studies are performed using as base model $b_{\theta^B}$ a CNN inspired by the popular AlexNet model~\cite{AlexNet}. It differs from AlexNet in that it uses 3 convolutional layers (instead of 5) and 1 fully connected layer (instead of 3) to work on the smaller-scale CIFAR100 dataset. We call this smaller network AlexNet-C100. The full specifications of the architecture are given in the supplementary material, including details about the learning policy. \LTR{The full specification of the architecture is as follows: {\footnotesize [CONV:1$\times$32$\times$5],[CONV:1$\times$32$\times$5],[CONV:1$\times$64$\times$5], [FC:100],[SM:100]} where {\footnotesize[CONV: 1$\times$32$\times$5]} 
denotes 1 convolutional layer containing 32 filters of size $5\times5$,  {\footnotesize[FC:100]} is a fully connected layer with 100 output units and  {\footnotesize[SM:100]} is a softmax layer that renormalizes the input to define a probability distribution over 100 classes (note that a {\footnotesize SM} layer does not contain parameters). Further details about the learning policy and the model can be found in the supplementary material.}

Our generalist is identical to this architecture with the only difference being that we set the number of units in the {\footnotesize FC} and {\footnotesize SM} layers to $K$, the number of specialties. 
The training is done from scratch. The learning alternates between updating network parameters $\theta^G$ and specialty labels $\ell$. The specialty labels are updated every 1 epoch of backpropagation over $\theta^G$. We use a random subset $\mathcal{S}$ of 10,000 images to build the confusion matrix. In the supplementary material we show some of the specialties learned by our generalist. Most specialties define intuitive clusters of classes, such as categories that are semantically or visually similar to each other (e.g., dolphin, seal, shark, turtle, whale).

Once the generalist is learned, we remove its {\footnotesize FC} layer and connect $K$ branches, each consisting of: {\footnotesize [CONV:1$\times$64$\times$5],[FC:c]} where {\footnotesize[CONV: 1$\times$64$\times$5] denotes 1 convolutional layer containing 64 filters of size $5\times5$,  {\footnotesize[FC:c]} is a fully connected layer with $c$ output units corresponding to the classes in the specialty (note that $c$ may vary from specialty to specialty)}. We link the $K$ FC layers of the branches to a {\em global} softmax over all $C$ classes (without parameters). The weights of each branch are randomly initialized. The full \nofe is trained via backpropagation using the same learning rate policy as for the base model. 

\subsubsection{Number of Experts and Specialty Balance.} The degree of specialization of the experts in our model is controlled by parameter $K$. Here we study how the value of this hyperapameter affects the final accuracy of the network. Furthermore, we also assess the importance of balancing the size of the specialties in connection with the value of $K$, since these two factors are interdependent. The method {\tt fully-balanced} (introduced in section~\ref{sec:generalist}) constrains all specialties to have equal size ($C/K$), while {\tt elasso} encourages softly the constraint over the size of specialties. The behavior of {\tt elasso} is defined by hyperparameter $\lambda$ which trades off the importance between having balanced specialties and good categorization accuracy over them. 

Table~\ref{table:accuracy_cifar100_alexfull_kclust} summarizes the recognition performance of our \nofe for different values of $K$ and the two ways of balancing specialty sizes. For {\tt elasso} we report accuracy using $\lambda=1000$, which was the best value according to our evaluation. We can immediately see that the two balancing methods produce similar recognition performance, with {\tt fully-balanced} being slightly better than {\tt elasso}. Perhaps surprisingly, {\tt elasso}, which gives the freedom of learning specialties of unequal size, is overall slightly worse than {\tt fully-balanced}. From this table we can also evince that our network of experts is fairly robust to the choice of $K$. As $K$ is increased the accuracy of each balancing method produces an approximate ``inverted U'' curve with the lowest performance at the two ends ($K=2$ and $K=50$) and the best accuracy for $K=5$ or $K=10$. Finally, note that all instantiations of our network of experts in this table achieve higher accuracy than the ``flat'' base model, with the exception of the models using $K=2$ experts which provide performance comparable to the baseline. 
Our best model ($K=10$ using {\tt fully-balanced}) yields a substantial improvement over the base model ($56.2\%$ versus $54.0\%$) corresponding to a relative gain in accuracy of about $4\%$. 

Based on the results of Table~\ref{table:accuracy_cifar100_alexfull_kclust}, all our subsequent studies are based on a \nofe architecture using $K=10$ experts and {\tt fully-balanced} for balancing.

%

%
\setlength{\tabcolsep}{4pt}
\begin{table}[t!]
\begin{center}
\caption{Top-1 accuracy ($\%$) on CIFAR100 for the base model (AlexNet-C100) and Network of Experts (\nofe) using varying number of experts ($K$) and two different specialty balancing methods. The best \nofe outperforms the base model by $2.2\%$ and all \nofe using $K>2$ experts yield better accuracy than the flat architecture.}
\label{table:accuracy_cifar100_alexfull_kclust}
{\scriptsize
\begin{tabular}{| c | c || c | c |c |c|c|}
\hline\noalign{\smallskip}
\bf Model & \bf Balancing Method & \bf K=2  &\bf  K=5 & \bf K=10 & \bf K=20 & \bf K=50 \\
\noalign{\smallskip}
\hline
\noalign{\smallskip}
\hline
\multirow{2}{*}{{\it \nofe}} & {\tt fully-balanced} & 53.3 & 55.0 & \bf 56.2 & 55.7 &  55.33\\
 & {\tt elasso}  $\lambda$=1000  & 53.93 &  53.6 & 55.59  &  55.3 & 55.3\\
\hline
{\smallskip}
Base: AlexNet-C100 & n/a & \multicolumn{5}{c|}{54.0}\\ 
\hline
\end{tabular}
}
\end{center}
\end{table}
\setlength{\tabcolsep}{1.4pt}

\LTR{
\subsubsection{Depths vs specialization.} The Network of Expert  assessed in the previous experiment has a total depth of 5 learnable layers (3 {\footnotesize CONV} layers in the trunk, and 1 {\footnotesize CONV} layer plus 1 {\footnotesize FC} layer in each branch). On the other hand, the base model has a depth of 4 learnable layers. As increased depth has been shown in some cases to lead to better accuracy~\cite{SimonyanZisserman:ICLR2015}, one could argue that the performance improvement in our \nofe comes merely from the additional convolutional layer. To disprove this, we retrained a variant of the base model augmented with an additional convolutional layer (with 64 filters) before the fully connected layer (thus producing a total depth of 5). 
We found that the additional layer leads to a degradation rather than an improvement in performance for the base network: the accuracy drops from $54.0\%$ to $51.3\%$. This suggests that the gain in accuracy derives from the tree-structure of our model and from the specialization performed by the experts rather than from mere additional depth or increased number of parameters. }

%

\LTR{
\subsubsection{Finetuning from the base model.} Our \nofe performs learning from scratch and does not require the learning of the base model in order to train the generalist and the experts. This is advantageous as it leads to a faster and more streamlined training procedure. However, one may wonder if learning the \nofe from the pretrained base model may actually lead to better performance. We attempted this experiment by finetuning the generalist from the base model (AlexNet-C100). The resulting generalist was then used to train the full \nofe, as usual. The accuracy achieved with this setup is $55.5\%$, thus inferior to the $56.2\%$ produced when learning from scratch. This suggests that the features learned by the base model by solving the hard classification over $C$ classes are providing a poor initialization for the generalist and the subsequent expert branches, which is instead the approach used in prior expert-based networks~\cite{HintonEtAl:arXivDISTILL,FarleyEtAl:ICLR2015,HD-CNNICCV15}.}

\subsubsection{Defining Specialties with Other Schemes.} Here we are study how the definition of specialties affects the final performance of the \nofe. Prior work~\cite{HintonEtAl:arXivDISTILL,FarleyEtAl:ICLR2015,HD-CNNICCV15} has proposed to learn groupings of classes by first training a CNN over all $C$ classes and then performing spectral clustering~\cite{AndrewNg:NIPS2011} of the confusion matrix. We tried this procedure by building the confusion matrix using the predictions of the base model (AlexNet-C100) over the entire CIFAR100 training set. We then partitioned the $C=100$ classes into $K=10$ clusters using spectral clustering. 
We learned a generalist optimized to categorize these $K$ clusters (without any update of the specialty labels) and then a complete \nofe. The performance of the resulting \nofe is illustrated in the second row of Table~\ref{table:other_startegies}. The accuracy is considerably lower than when learning specialties with our approach (first row). The third row shows accuracy achieved when training the generalist and subsequently the full \nofe on a random partitioning of the classes into $K=10$ clusters of equal size. The performance is again inferior to that achieved with our approach. Yet, surprisingly it is better than the accuracy produced by spectral clustering. We believe that this happens because spectral clustering yields highly imbalanced clusters that lead to poor performance of the \nofe. 


\begin{table}[t!]
\parbox{.41\linewidth}{
\centering
\caption{CIFAR100 top-1 accuracy of \nofe models trained on different definitions of specialties: ours, spectral clusters of classes and random specialties of equal size. }
\label{table:other_startegies}
{\scriptsize
\begin{tabular}{| l || c | c |c |}
\hline\noalign{\smallskip}
 \bf  Specialty Method & Accuracy \%\\
\noalign{\smallskip}
\hline
\noalign{\smallskip}
{Our Method (joint training)} & \bf 56.2 \\
\hline
{Spectral Clustering} &  53.2\\
\hline
{Random Balanced Specialties} & 53.7 \\
 \hline
\end{tabular}
}
}
\hfill
\parbox{.54\linewidth}{
\centering
\caption{CIFAR100 top-1 accuracy ($\%$) for 5 different CNN base architectures and corresponding \nofe models. In each of the five cases our \nofe yields improved performance.}
\label{table:accuracy_cifar100_allarchs}
{\scriptsize
\begin{tabular}{| l || c | c |}
\hline\noalign{\smallskip}
 \bf  Architecture & Base Model & \nofe \\
\noalign{\smallskip}
\hline
\noalign{\smallskip}
AlexNet-C100 & 54.04 & \bf 56.24\\
\hline
AlexNet-Quick-C100 & 37.94 & \bf 45.58\\
\hline
VGG11-C100 & 68.48 & \bf 69.27\\
\hline
NIN-C100 & 65.41  & \bf 67.96\\
\hline
ResNet56-C100 & 73.52 & \bf 76.24\\    
\hline
\end{tabular}
}
}
\end{table}

\subsubsection{Varying the Base CNN.} The experiments above were based on AlexNet-C100 as the base model. Here we study the generality of our approach by considering 4 other base CNNs (see supplementary material for architecture details):

\noindent 1) {\em AlexNet-Quick-C100.} This base model is a slightly modified version of AlexNet-C100 where we added an extra FC layer (with 64 output units) before the existing FC layer and removed local response normalization. This leads to much faster convergence but lower accuracy compared to AlexNet-C100. 
After training the generalist, we discard the two FC layers and attach $K$ branches, each with the following architecture: {\footnotesize [CONV:1$\times$64$\times$5], [FC:64],[FC:10]}. 

\noindent 2) {\em VGG11-C100.} This model is inspired by the VGG11 architecture described in~\cite{SimonyanZisserman:ICLR2015} but it is a reduced version to work on the smaller-scale CIFAR100. We take this model from~\cite{caffe}. \LTR{The base model is: {\footnotesize[CONV:2$\times$64$\times$3], [CONV:2$\times$128$\times$3],[CONV: 4$\times$256$\times$3], [FC:1024], [FC:1024], [FC:100]}. }
In the expert branch we use 2 convolutional layers and 3 FC layers 
to match the number of FC layers in the base model but we scale down the number of units to account for the multiple branches. The branch architecture is: {\footnotesize[CONV:2$\times$256$\times$3],[FC:512],[FC:512],[FC:10]}. 

\noindent 3) {\em NIN-C100.} This is a ``Network-In-Network'' architecture~\cite{NIN} that was used in~\cite{HD-CNNICCV15} as base model to train the hierarchical HD-CNN network on CIFAR100. 

\noindent 4) {\em ResNet56-C100.} This is a residual learning network~\cite{HeEtAl:arXivResNet} of depth 56 using residual blocks of 2 convolutional layers. We modeled it after the ResNet-56 that the authors trained on CIFAR10. To account for the 10X number of classes in CIFAR100 we quadruple the number of filters in each convolutional layer. 
\LTR{The supplementary material includes the full specification of this network. }
To maintain the architecture homogenous, each expert branch consists of a residual block (rather than a CONV layer) followed by average pooling and an FC layer.

These 4 \nofe models were trained using $K=10$ and {\tt fully-balanced}. The complete results for these 4 base models and their derived \nofe are given in Table~\ref{table:accuracy_cifar100_allarchs} (for completeness we also include results for the base model AlexNet-C100, previously considered). In every case, our \nofe achieves higher accuracy than the corresponding base model. In the case of AlexNet-Quick-C100, the {\em relative} improvement is a remarkable $20.1\%$. NIN-C100 was used in~\cite{HD-CNNICCV15} as base model to train the hierarchical HD-CNN. The best HD-CNN network from~\cite{HD-CNNICCV15} gives $67.38\%$, whereas we achieve $67.96\%$. Furthermore, our \nofe is twice as fast at inference (0.0071 vs 0.0147 secs) and it has about half the number of parameters (4.7M vs 9.2M).
Finally, according to the online listing of top results on CIFAR100~\cite{rodrigourl} at the time of this submission, the accuracy of $76.24\%$ obtained by our \nofe built from ResNet56-C100 is the best result ever achieved on this benchmark (the best published accuracy is $72.60\%$~\cite{SnoekEtAl:ICML2015} and the best result considering also unpublished work is $75.72\%$~\cite{ELU}).

\begin{table}[t!]
\centering
\caption{We add layers to the base models in order to match the number of parameters of our \nofe models (the last column reports results with base networks having both the same depth and the same number of parameters as our models). The table reports accuracy ($\%$) on CIFAR100. This study suggests that the accuracy gain of our \nofe does not derive from an increase in depth or number of parameters, but rather from the specialization performed by branches trained on different subsets of classes.}
\label{table:depthnparamsVSspecialization}
{\scriptsize
\begin{tabular}{| l || c | c || c | c |}
\hline\noalign{\smallskip}
 \bf  Architecture &  \shortstack{Original\\base model} & \shortstack{\nofe\\{ }} & \shortstack{Modified base model\\matching \nofe  \# params} & \shortstack{Modified base model matching\\\nofe \# params {AND} depth}
 \\
\noalign{\smallskip}
\hline
\noalign{\smallskip}
AlexNet-C100 & 54.04 & \bf  56.24 & 30.75 & 50.21 \\  
\hline
VGG11-C100 & 68.48 & \bf 69.27 & 68.68 & 68.21 \\
\hline
ResNet56-C100 & 73.52 & \bf 76.24 & 73.50  & 73.88 \\    
\hline
\end{tabular}
}
\end{table}

\subsubsection{Depth and \# parameters vs. specialization.} Our \nofe differs from the base model in total depth (i.e., number of nonlinearities to compute the output) as well as in number of parameters, because of the additional layers (or residual blocks) in the branches. Here we demonstrate that the improvement does not come from the increased depth or the different number of parameters, but rather from the novel architecture and the procedure to train the experts. To show this, we modify the base networks to match the number of parameters of our \nofe models. We consider two ways of modifying the base models. In the first case, we add to the original base network $K=10$ times the number of convolutional layers (or residual blocks) contained in one branch of our \nofe (since we have $K=10$ branches, each with its own parameters). This produces base networks matching the number of parameters of our \nofe models but being deeper. The other solution is to add to the base model a number of  layers (or residual blocks) equal to the number of such layers in one branch of the \nofe, but to increase the number of convolutional filters in each layer to match the number of parameters. This yields modified base networks matching both the total depth and the number of parameters of our models. Table~\ref{table:depthnparamsVSspecialization} reports the results for 3 distinct base models. The results show unequivocally that our \nofe models outperform base networks of equal learning capacity, even those having same depth.

\subsubsection{Finetuning \nofe from generalist vs. learning from scratch} Here we want to show that in addition to defining good specialties, the learning of the generalist provides a beneficial pretraining of the trunk of our \nofe. To demonstrate this, we trained \nofe models from scratch (i.e., random weights) using the specialties learned by the generalist. This training setup yields an accuracy of $49.53\%$ when using AlexNet-C100 as base model and $73.95\%$ when using ResNet56-C100. Note that learning the \nofe models from the pretrained generalist yields much better results: $56.2\%$ for AlexNet-C100 and $76.24\%$ for ResNet56-C100. This suggests that the pretraining of the trunk with the generalist is crucial.

\LTR{We have also tried to see if learning the \nofe from a pretrained base model may actually lead to better performance. We attempted this experiment by finetuning the generalist from the base model (AlexNet-C100). The resulting generalist was then used to train the full \nofe, as usual. The accuracy achieved with this setup is $55.5\%$, thus inferior to the $56.2\%$ produced when learning the generalist from scratch. This suggests that the features learned by the base model by solving the hard classification over $C$ classes are providing a poor initialization for the generalist and the subsequent expert branches, which is instead the approach used in prior expert-based networks~\cite{HintonEtAl:arXivDISTILL,FarleyEtAl:ICLR2015,HD-CNNICCV15}.}

\subsection{Categorization on ImageNet}

In this subsection we evaluate our approach on the ImageNet 2012 classification dataset~\cite{ImageNet}, which includes images of 1000 object categories. The training set consists of 1.28M photos, while the validation set contains 50K images. We train on the training set and use the validation set to assess performance. 

Our base model here is the Caffe~\cite{caffe} implementation of AlexNet~\cite{AlexNet}. It contains 8 learned layers, 5 convolutional and 3 fully connected. As usual, we first train the generalist by simply changing the number of output units in the last {\footnotesize FC} layer to $K$, using our joint learning over network weights and specialty labels. We experimented with two variants: one generalist using $K=10$ experts, and one with $K=40$. Both were trained using {\tt fully-balanced} for specialty balancing. The complete \nofe is obtained from the generalist by removing the 3 {\footnotesize FC} layers and by connecting the last {\footnotesize CONV} layer to $K$ branches, each with architecture: {\footnotesize [CONV:1$\times$256$\times$3],[FC:1024],[FC:1024],[FC:100]}. Note that while the base model (and the generalist) use layers of dimensionality 4096 for the first two {\footnotesize FC} layers, the expert branches use 1024 units in these layers to account for the fact that we have $K=10$ parallel branches. 

\begin{figure}[t!]
\centering
\begin{minipage}[t!]{.55\textwidth}
\centering
\includegraphics[width=6.5cm]{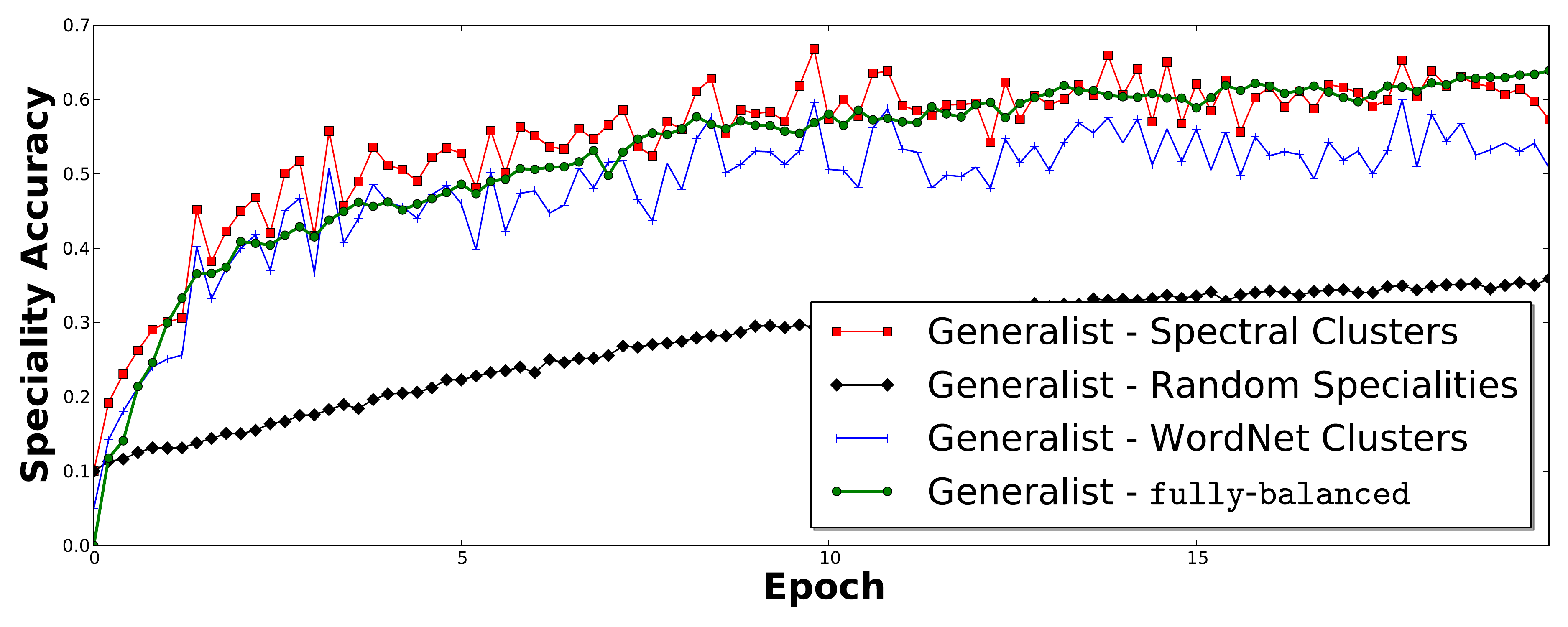}
\caption{Generalist accuracy for different definitions of specialties (ours, spectral clusters, random specialties and WordNet clusters). The accuracy is assessed on the validation set for varying training iterations of the generalist.}
\label{fig:specialtyacc}
\end{minipage}\hfill
\begin{minipage}[t!]{.4\textwidth}
\centering
\captionof{table}{Top-1 accuracy on the ImageNet validation set using AlexNet and our \nofe.}
\label{table:accuracy_imagenet_alexnet_val}
{\scriptsize
\begin{tabular}{| l || l|c| c |c|}
\hline\noalign{\smallskip}
 \bf  Approach & Top-1 \% & \# params\\
\noalign{\smallskip}
\hline
\noalign{\smallskip}
Base: AlexNet-Caffe & 58.71 & $60.9M$\\
\hline
\hline
\nofe, K=10\\
{\tt fully-balanced} & \bf 61.29 &$40.4M$ \\
\hline
\nofe, K=40\\
{\tt fully-balanced} & 60.85 &$151.4M$ \\
\hline
\hline
\nofe, K=10\\spectral clustering  & 56.10 & $40.4M$ \\
\hline
\end{tabular}
}
\bigbreak
\bigbreak
\end{minipage}
\end{figure}

We also tested two other ways to define specialties: 1) spectral clustering on the confusion matrix of the base model (on the validation set) and 2) WordNet clustering. The WordNet specialties are obtained by ``slicing'' the WordNet hierarchy at a depth that intersects $K=10$ branches of this semantic tree and then aggregating the ImageNet classes of each branch into a different specialty. We also trained a generalist on a random balanced partition of the 1000 classes. Figure~\ref{fig:specialtyacc} shows the classification accuracy of the generalist for these different ways of defining specialties. The accuracy is assessed on the validation set and measures how accurately the generalist recognizes  the $K=10$ specialties as a function of training epochs. We can see that while the CNN trained on the fixed spectral clusters does best in the initial iterations, our generalist (with specialty labels updated every fifth of an epoch) eventually catches up and matches the performance of the spectral generalist. The generalist trained on WordNet clusters and the one trained on random specialties do much worse.

We then built \nofe models from the spectral generalist and our own generalist (we did not train \nofe models from the random or WordNet generalists due to their poor performance). Table~\ref{table:accuracy_imagenet_alexnet_val} shows the accuracy of all these models on the validation set. For each we report top-1 accuracy (averaging over 10 crops per image, taken from the 4 corners and the center, plus mirroring of all of them). We also include results for our approach using $K=40$ experts. It can be seen that our \nofe with $K=10$ experts outperforms the base model, yielding a relative improvement of $4.4\%$. Instead, the \nofe trained on spectral clusters does worse than the base model, despite the good accuracy of the spectral generalist as noted in Fig.~\ref{fig:specialtyacc}. We believe that this happens because of the large imbalance of the spectral specialties, which cause certain experts to have classification problems with many more classes than others. Our \nofe with $K=40$ experts does worse than the one with $K=10$ experts, possibly because of excessive specialization of the experts or overfitting (see number of parameters in last column). Note that our \nofe with $K=10$ experts has actually fewer parameters than the base model and yet it outperforms it by a good margin. This indicates that the improvement comes from the structure of our network and the specialization of the experts rather than by larger learning capacity. 

In terms of training time, the base model required 7 days on a single NVIDIA K40 GPU. The \nofe with $K=10$ experts took about 12 days on the same hardware (2 days for the generalist and 10 days for the training of the full model). On CIFAR100 the ratio of the training time between \nofe and base models was about 1.5X (0.5X for the generalist, 1X for the full model). Thus, there is an added computational cost in training our architecture but it is fractional.

Finally, we also evaluated the base model and our \nofe of $K=10$ experts on the 100K test images of ImageNet, using the test server. The base model achieves a top-1 accuracy of $58.83\%$ while our \nofe yields a recognition rate of $61.48\%$, thus confirming the improvements seen so far also on this benchmark.

\section{Conclusions}

In this paper we presented a novel approach that decomposes large multi-class classification into the problem of learning 1) a generalist network distinguishing coarse groupings of classes, called specialties, and 2) a set of expert networks, each devoted to recognize the classes within a specialty. Crucially, our approach learns the specialties and the generalist that recognizes them {\em jointly}. Furthermore, our approach gives rise to a single tree-structured model that is fine-tuned over the end-objective of recognizing the original set of classes. We demonstrated the generality of our approach by adapting several popular CNNs for image categorization into networks of experts. In each case this translated into an improvement in accuracy at very little added training cost. Software implementing our method and several pretrained \nofe models are available at \url{http://vlg.cs.dartmouth.edu/projects/nofe/}.

\subsubsection{Acknowledgements.} We are grateful to Piotr Teterwak for assisting with an early version of this project. We thank Du Tran and Qiang Liu for helpful discussions. This work was funded in part by NSF CAREER award IIS-0952943 and NSF award CNS-1205521. We gratefully acknowledge NVIDIA for the donation of GPUs used for portions of this work.

\clearpage

\newpage 

\appendix

\section{Supplementary Material}

This supplementary material is organized as follows: in subsection~\ref{nn_section} we discuss a simple Nearest Neighbor (NN) retrieval experiment aimed at illustrating the differences between the features learned by our \nofe and the base model; in subsection~\ref{clusters_section} we illustrate some examples of specialties learned from CIFAR100 and ImageNet; in subsection~\ref{Finetuning_from_the_base_model} we demonstrate that finetuning our \nofe from the generalist produces better results than finetuning from a pretrained base model; in subsection~\ref{depth_section} we perform a ``depth vs specialization'' analysis on the large-scale ImageNet dataset; in subsection~\ref{specialty_sizes} we visualize distributions of specialty sizes obtained by varying the regularization hyperparameter of {\tt elasso}; in subsection~\ref{spec_section} we present a complete specification of all network models presented in the paper together with details about the training procedure; we conclude with subsection~\ref{software_section} where we discuss our software implementation.

\subsection{Nearest Neighbor Retrieval}
\label{nn_section}

\begin{figure}
\centering
\includegraphics[scale=0.6]{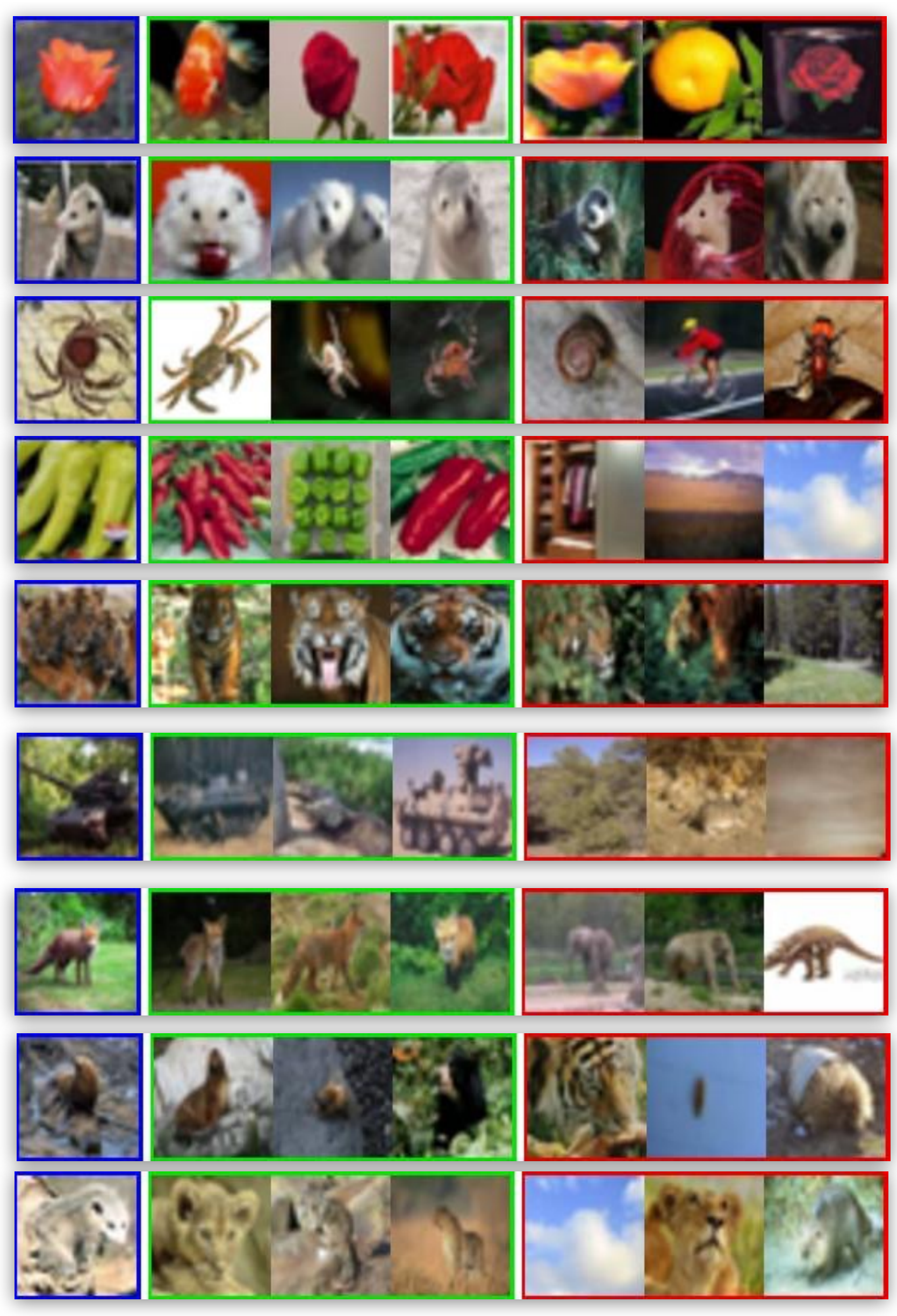}
\caption{Nearest Neighbor (NN) retrieval on CIFAR100.  The first column shows the query image. The first set of 3 images represents NNs retrieved using our expert features. The second set of 3 images represents NNs obtained with features from the base model.}
\label{fig:NNCIFAR100}
\end{figure}

Figure~\ref{fig:NNCIFAR100} shows a simple Nearest Neighbor (NN) retrieval experiment on CIFAR100 using features from our experts vs the base model, for the architecture AlexNet-C100. For each query image (first column of the Figure) we perform a NN search in the CIFAR100 test set using as feature representation the last {\footnotesize CONV} layer of the expert corresponding to the predicted class of the query. The first block of 3 images near the query shows the 3 NNs retrieved in this fashion. The second set of 3 images represents NNs obtained the using as features the last {\footnotesize CONV} layer of the base model. It can be seen that in many cases the images retrieved with the expert features match the class of the query, while the base model features yield many mismatches.

\begin{figure}
\centering
\includegraphics[scale=0.65]{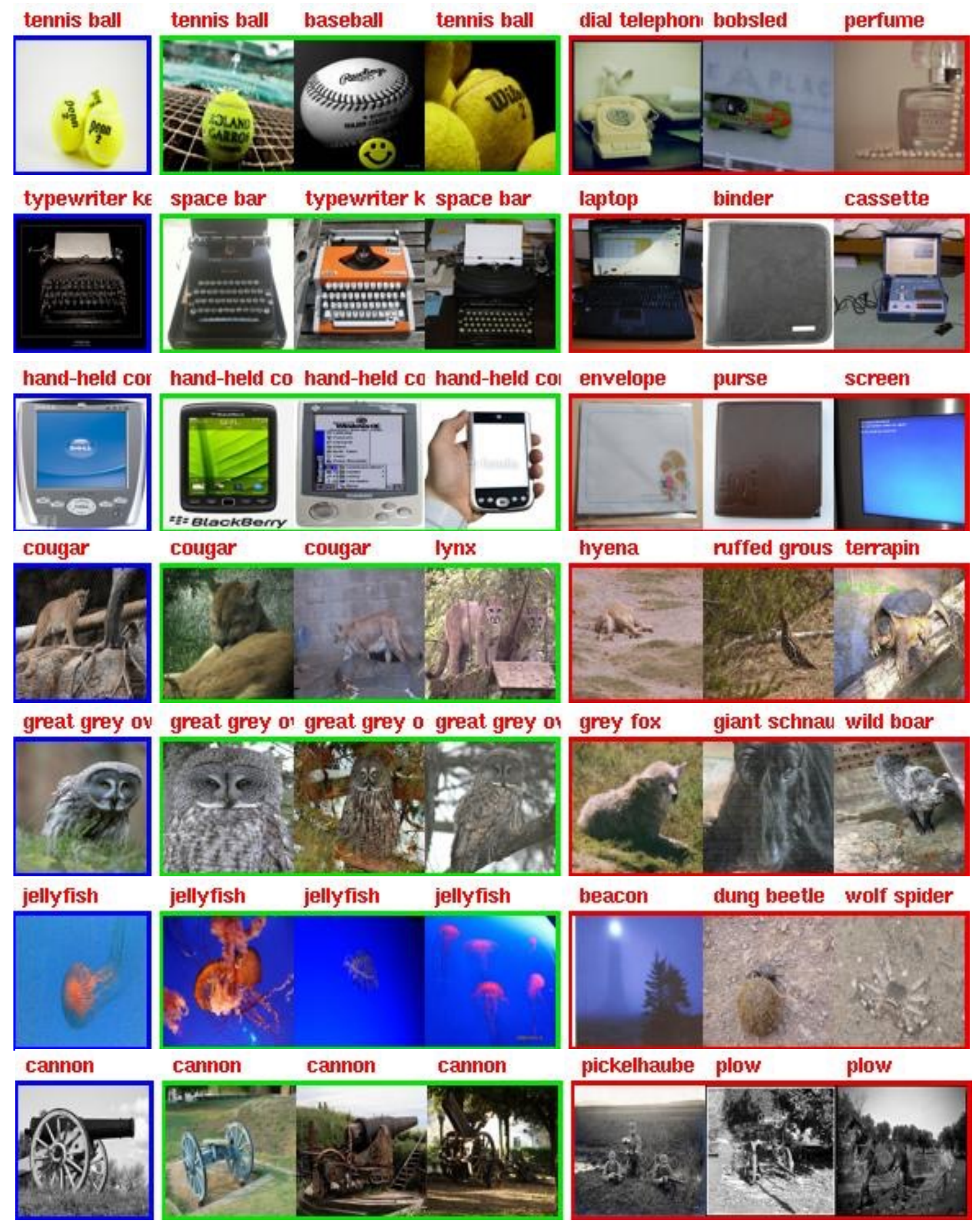}
\caption{Nearest Neighbor (NN) search on ImageNet. The first column shows the query image. The first set of 3 images includes NNs retrieved using as features the activations from the first FC layer of the expert branch associated to the predicted class of the query. The second block of 3 images represents NNs obtained using the first FC layer of the base model. Above each image we report its ImageNet category. Our expert features yield NNs that are semantically related to the query.}
\label{fig:NNImageNet}
\end{figure}

Figure~\ref{fig:NNImageNet} shows selected retrieval results on the large-scale ImageNet dataset. The first set of 3 images shows NNs obtained using as features the activations from the first FC layer of the expert branch corresponding to the predicted class of the query. The second block of 3 images represents NNs obtained using the first FC layer of the base model. Above each image we report its ImageNet category. We see that the NNs obtained with expert features include many true positives (images of the same class as the query) or, when making mistakes, images of related classes (e.g., a picture of a lynx is retrieved for a cougar query).

\subsection{Learned Specialties}
\label{clusters_section}

In Table~\ref{table:clusters_k10_alex_cifar} we show some of the specialties learned by our generalist on CIFAR100 when using $K=10$ and AlexNet-C100 as architecture. Most classes within each specialty are semantically or visually similar to each other, but some outliers are also present (e.g., the class ``skyscraper'' in the first specialty, which otherwise contains natural classes). Table~\ref{table:clusters_k20_alex_cifar} shows that when using $K=20$ we obtain specialties that are more homogeneous and that make more intuitive sense, since the generalist can perform a finer subdivision of the original $C=100$ categories. In terms of final categorization accuracy, we have seen that the \nofe with $K=20$ does not provide better results compared to $K=10$. This is most likely due to the higher number of parameters in the configuration with $K=20$, which may cause overfitting. 

In Table~\ref{table:clusters_k10_alex_imagenet}, we show some of the specialists learned by our generalist on ImageNet, using $K=10$ and AlexNet as architecture.

\setlength{\tabcolsep}{4pt}
\begin{table}[t!]
\begin{center}
\caption{A few specialties learned by our generalist from CIFAR100 using $K=10$ with AlexNet-C100 as architecture.}
\label{table:clusters_k10_alex_cifar}
{\scriptsize
\begin{tabular}{| c | }
\hline 
caterpillar, lawn\_mower,  crab, lizard,  forest, maple\_tree, oak\_tree, pine\_tree, willow\_tree, skyscraper \\
\hline
\hline 
 pickup\_truck, road, streetcar, tank, tractor, train, bus, aquarium\_fish,can, house \\
\hline
\hline 
hamster, leopard, lion, tiger, possum, rabbit, raccoon,  squirrel, snail, wolf \\
\hline
\hline 
bicycle, motorcycle, bottle, camel, cattle, elephant, fox, kangaroo, seal, trout \\

\hline

\end{tabular}
}
\end{center}
\end{table}

\setlength{\tabcolsep}{4pt}
\begin{table}[t!]
\begin{center}
\caption{Some example specialties learned from CIFAR100 using $K=20$ with AlexNet-C100 as architecture. Note how, compared to the case for $K=10$ shown in Table~\ref{table:clusters_k10_alex_cifar}, here the finer subdivision of classes yields specialties that are more homogeneous and that include categories that are more closely related. }
\label{table:clusters_k20_alex_cifar}
{\scriptsize
\begin{tabular}{| c | }
\hline 
 maple\_tree, oak\_tree, pine\_tree, willow\_tree, palm\_tree  \\
\hline
\hline 
apple, cloud, poppy, rose, tulip \\
\hline
\hline 
dolphin, seal, shark, turtle, whale \\
\hline
\hline 
baby, boy, girl, man, woman \\
\hline 

\end{tabular}
}
\end{center}
\end{table}
\setlength{\tabcolsep}{1.4pt}

\setlength{\tabcolsep}{4pt}
\begin{table}[t!]
\begin{center}
\caption{A few specialties learned by our generalist from ImageNet using AlexNet as architecture and $K=10$.}
\label{table:clusters_k10_alex_imagenet}
{\scriptsize
\begin{tabular}{| p{10cm}| }
\hline 
  
can opener, tinopener, carpenter's kit, cassette, cassette player, cellular phone, hand-held computer, iPod, joystick, speaker, computer mouse, parking meter, pay-phone, pay-station, photo copier, polaroid camera, printer, remote control, scale, stove, electric switch, . . .\\ 
\hline \hline 
cock, hen, black grouse, macaw, drake, flamingo, great pyrenees, standard poodle, ice bear, polar bear, Arabian camel, baseball, golfball, lifeboat, pencil sharpener, schoolbus, streetcar, tram, trolleycar, trolley bus, trolley coach, lemon, rapeseed, yellow lady-slipper, rosehip, coralfungus, agaric, . . . \\ 
\hline \hline 

hyena, zebra, Indian elephant,  African elephant, lion fish, spiny lobster, seacrawfish, cray fish, rock lobster, chain, chain mail, ring mail, necklace, modem, packet, puck, hockeypuck, comicbook, crossword puzzle, menu, . . .\\ 
\hline \hline 
Bordercollie, GreaterSwissMountaindog, Bernesemountaindog, Appenzeller, EntleBucher, affenpinscher, monkeypinscher, monkeydog, West Highland white terrier,  white wolf, Arctic wolf,  Arctic fox, white fox, Americanblackbear, sorrel, ox ,water buffalo, water ox, Asiatic buffalo, bison, bearskin, busby, shako, guillotine, harp, horse-cart, hourglass, megalith, oxcart, snowplow, thatch, . . .\\ \hline 
\end{tabular}
}
\end{center}
\end{table}
\setlength{\tabcolsep}{1.4pt}


\subsection{Finetuning from the base model} 
\label{Finetuning_from_the_base_model}

Our \nofe performs learning from scratch and does not require the learning of the base model in order to train the generalist and the experts. This is advantageous as it leads to a faster and more streamlined training procedure. However, one may wonder if learning the \nofe from the pretrained base model may actually lead to better performance. We attempted this experiment by finetuning the generalist from the base model (AlexNet-C100). The resulting generalist was then used to train the full \nofe, as usual. The accuracy achieved with this setup is $55.5\%$, thus inferior to the $56.2\%$ produced when learning from scratch. This suggests that the features learned by the base model by solving the hard classification over $C$ classes are providing a poor initialization for the generalist and the subsequent expert branches, which is instead the approach used in prior expert-based networks~\cite{HintonEtAl:arXivDISTILL,FarleyEtAl:ICLR2015,HD-CNNICCV15}.


\subsection{Depth vs Specialization}
\label{depth_section}

Our \nofe typically has one convolutional layer more than the base model, as we include a convolutional layer in each branch. In the experimental section of our paper we discussed the results of a CIFAR100 experiment, which shows that the improvement in accuracy achieved by our approach is not due to the increased depth of the model but rather from its structure. Here we report the same experiment but for the case of ImageNet: we trained a variant of the AlexNet-Caffe base model that has one additional convolutional layer ({\footnotesize CONV:1$\times$256$\times$3}) such that this network has total depth exactly equal to the depth of the \nofe presented in Table~5 of our paper. We found that this deeper variant of the base model yields an accuracy of $56.91\%$, thus lower than the shallower base network and much lower than the accuracy of $61.29\%$ achieved by our \nofe. This indicates once again that the critical improvement in our approach comes from the specialization performed by the experts rather than from the additional layer.


\subsection{Distribution of specialty sizes}
\label{specialty_sizes}

\begin{figure}[t!]
\centering
\includegraphics[width=10cm]{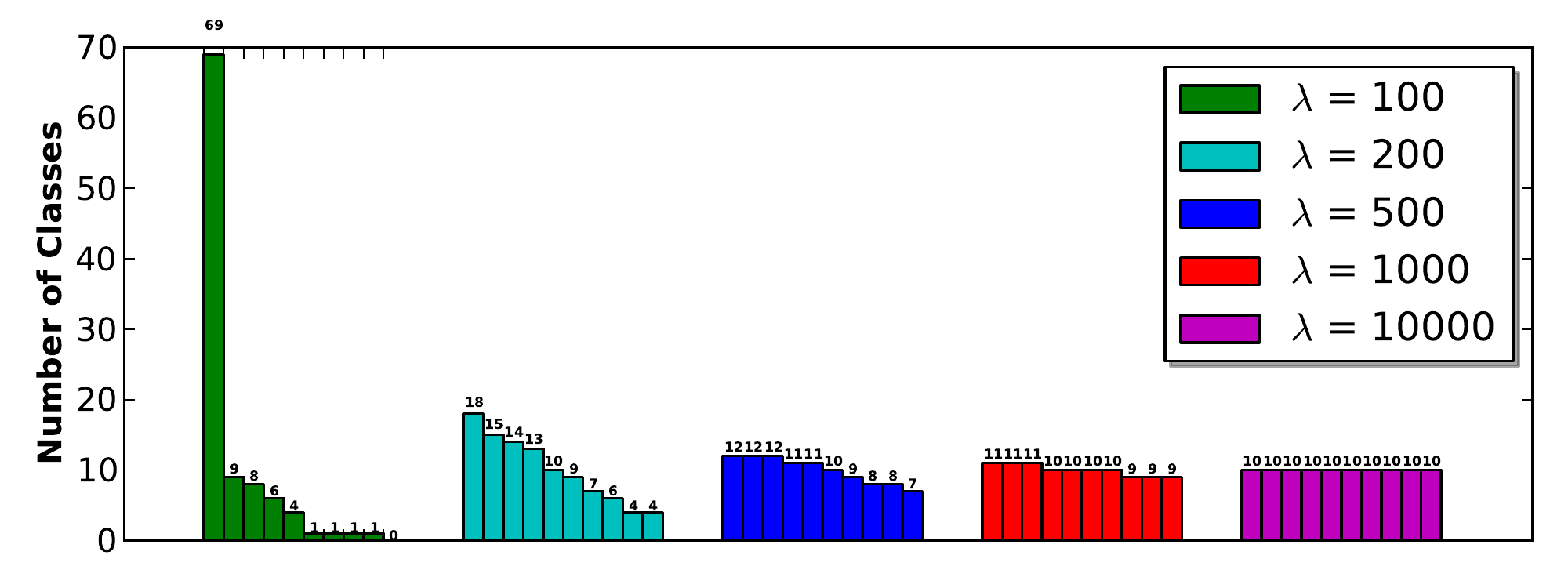}
\caption{Distribution of specialty sizes learned by {\tt elasso} for different $\lambda$ values (100, 200, 500, 1000, and 10000) using $K=10$ experts. For each $\lambda$ the specialties are sorted in decreasing size. Small values of $\lambda$ yield highly imbalanced specialties, while large values of $\lambda$ force specialties to have nearly equal size.}
\label{fig:cifar100_elasso_distribution_per_Lambda}
\end{figure}

In the paper we demonstrated the importance of balancing the size of the specialties. One of the methods investigated is {\tt elasso}, which encourages softly the constraint over the size of specialties. The behavior of {\tt elasso} is defined by hyperparameter $\lambda$ which trades off the importance between having balanced specialties and good categorization accuracy over them. Figure~\ref{fig:cifar100_elasso_distribution_per_Lambda} shows the different distributions in specialty size obtained for different values of $\lambda$ with $K=10$ experts. Small values of $\lambda$ (e.g., 100 or 200) produce specialties that are highly uneven in size. Conversely, a large value of $\lambda$ yields perfectly balanced classes. 

\subsection{Network Specifications}
\label{spec_section}

In this subsection, we provide full specification of all networks used in our experiments on the CIFAR100 and ImageNet datasets. We also discuss the details of the learning policy for each model. 

Tables~\ref{table:alex_c100_config},~\ref{table:alex_quick_c100_config},~\ref{table:vgg_c100_config},~\ref{table:nin_c100_config}, and~\ref{table:res_c100_config} list the specifications of the 4 architectures used on CIFAR100: AlexNet-C100, AlexNet-Quick-C100, VGG11-C100, NIN-C100, and ResNet56-C100. Table~\ref{table:alex_imagenet_config} provides the details of the AlexNet-Caffe architecture used for the large-scale ImageNet experiment (Table~4 in the paper).

Each table lists the specification of a network in two columns. The first column provides the details of both the generalist and the base model, which always share the same architecture, except for the number of units in the last fully connected layer and the training policy. The second column shows the Network of Experts (\nofe) obtained by removing the last fully connected layer of the generalist and by attaching to it $K$ expert branches. The architecture of the branches is listed in the yellow blocks.

Each entry in the tables represents a stack of layers, i.e., a set of one or more layers (convolutional or fully connected) connected directly to each other, and usually followed by a pooling layer. In all architectures we used rectified linear units (ReLU). In the following, we illustrate the notation used to describe each layer type.
\subsection*{Notation:}
\begin{itemize}
\item Convolutional layer: \\
 {\footnotesize CONV: $<$number of layers$>\times<$number of filters$>\times<$filter size$>$}  \\
Example:  {\footnotesize CONV: 1$\times$32$\times$5}  means 1 convolutional layer with 32 filters of size 5$\times$5. \\

\item Pooling layer: \\
 {\footnotesize POOL: $<$filter size$>$,$<$stride size$>$,$<$pooling type$>$}  \\
Example:  {\footnotesize POOL: 3,2,Max}: maximum pooling layer of filter size 3 and stride 2. Two types are used: Max (Maximum), Ave (Average). \\

\item LRN: Local Response Normalization layer \\

\item Fully Connected layer: \\
 FC: $<$size$>$
 Example:  {\footnotesize FC:$C$} indicates a fully connected layer with $C$ output units. 
 We use $C$ to denote to number of classes and $K$ to indicate the number of specialties. For simplicity, we assume that the architecture of the \nofe  illustrated in all tables is based on the \texttt{fully-balanced} method to train the generalist. This implies that each speciality contains the same number of classes ($C/K$).\\

\item Residual block:\\
Residual blocks are used in Residual Networks~\cite{HeEtAl:arXivResNet}. We use big braces to indicate  residual blocks. For example, the following notation denotes a concatenation of $9$ residual blocks, each of which consists of a {\footnotesize CONV:1$\times$64$\times$3} layer and another {\footnotesize CONV:1$\times$64$\times$3} layer, with batch normalization and scaling layers in between, as proposed in~\cite{HeEtAl:arXivResNet}:
\[ 
\left \{
  {\footnotesize 
    \begin{tabular}{l}
  CONV: 1$\times$64$\times$3 \\
  CONV: 1$\times$64$\times$3
  \end{tabular}
    }
\right \}\times9
\]    
\\

\item Expert branch: \\
An expert branch is a stack of one or more layers. A \nofe includes $K$ parallel branches connected to the trunk. We highlight the expert branches with yellow color in each table, and denote the $i$-th expert branch with ``Expert$^{(i)}\rightarrow$ ". \\

\end{itemize}



\setlength{\tabcolsep}{4pt}
\begin{table}[]
\begin{center}
\caption{\bf AlexNet-C100 (trained on CIFAR100)}
\label{table:alex_c100_config}

\begin{tabular}{ |p{5.5cm}|p{5.5cm}|  }
\hline
\cellcolor{blue!25} \bf{ Generalist and Base Model} &  \cellcolor{red!25} \bf{ Network of Experts (\nofe)} \\ 
\hline
\hline
 \layerfont{CONV: 1$\times$32$\times$5 } & \layerfont {CONV: 1$\times$32$\times$5}  \\ 
\layerfont{POOL: 3,2,Max}  & \layerfont{POOL: 3,2,Max}    \\
\layerfont{LRN}  & \layerfont{LRN}  \\
\hline
\hline 
\layerfont{CONV: 1$\times$32$\times$5}  & \layerfont{CONV: 1$\times$32$\times$5}  \\ 
\layerfont{POOL: 3,2,Ave}  & \layerfont{POOL: 3,2,Ave}    \\
\layerfont{LRN}  & \layerfont{LRN}  \\
\hline
\hline 
\layerfont{CONV: 1$\times$64$\times$5}  & \layerfont{CONV: 1$\times$64$\times$5}  \\ 
\layerfont{POOL: 3,2,Ave}  & \layerfont{POOL: 3,2,Ave}    \\
						 & \layerfont{LRN}   \\
\hline 
\hline
 					 &  \expertlayerfont{Expert$^{(i)}\rightarrow$ CONV: 1$\times$64$\times$5}  \\ 
					 & \expertlayerfont{Expert$^{(i)}\rightarrow$ POOL: 3,2,Ave}  \\ 
\hline
\hline
\layerfont{\parbox[t]{3.5cm}{FC: $K$ (Generalist) \\ FC: $C$ (Base Model) }} 	           & \expertlayerfont{Expert$^{(i)}\rightarrow$ FC: $C/K$}  \\ 
\hline
\hline
\multicolumn{2}{|c|}{\bf Data preprocessing and augmentation} \\
\hline
\lrfont {
Input: $32\times32$ \\
Image mean subtraction \\
} & 
\lrfont {
Input: $32\times32$ \\
Image mean subtraction \\
}
\\
\hline
\multicolumn{2}{|c|}{\bf Learning Policy} \\
\hline
\lrfont {
\textit{Generalist:} \\ 
Learning rate: 0.001  (Fixed)  \\ 
0.001 : 60 Epochs  \\
Momentum: 0.9 \\
Weight decay: 0.004 \\ 
Weight initialization: Random \\
MiniBatch size: 100\\

\textit{Base Model:} \\ 
Learning rate: 0.001  (lowered twice)  \\ 
140 Epochs  \\
Momentum: 0.9 \\
Weight decay: 0.004 \\ 
Weight initialization: Random \\
MiniBatch size: 100 \\
 } 
 & 
 \lrfont {
 Learning rate: 0.001  till 0.00001 (lowered twice) \\
0.001 : 120 Epochs \\
0.0001 : 10 Epochs \\
0.00001 : 10 Epochs \\

Momentum: 0.9 \\
Weight decay: 0.004 \\
Weight initialization: Random \\
 MiniBatch size: 100 \\
 }

 \\
\hline 
\end{tabular}

\end{center}
\end{table}
\setlength{\tabcolsep}{1.4pt}

\setlength{\tabcolsep}{4pt}
\begin{table}[]
\begin{center}
\caption{\bf AlexNet-Quick-C100 (trained on CIFAR100)}
\label{table:alex_quick_c100_config}

\begin{tabular}{ |p{5.5cm}|p{5.5cm}|  }
\hline
\cellcolor{blue!25} \bf{ Generalist and Base Model} &  \cellcolor{red!25} \bf{ Network of Experts (\nofe)} \\ 
\hline
\hline
 \layerfont{CONV: 1$\times$32$\times$5 } & \layerfont {CONV: 1$\times$32$\times$5}  \\ 
\layerfont{POOL: 3,2,Max}  & \layerfont{POOL: 3,2,Max}    \\
\hline
\hline 
\layerfont{CONV: 1$\times$32$\times$5}  & \layerfont{CONV: 1$\times$32$\times$5}  \\ 
\layerfont{POOL: 3,2,Ave}  & \layerfont{POOL: 3,2,Ave}    \\
\hline
\hline 
\layerfont{CONV: 1$\times$64$\times$5}  & \layerfont{CONV: 1$\times$64$\times$5}  \\ 
\layerfont{POOL: 3,2,Ave}  & \layerfont{POOL: 3,2,Ave}    \\
\hline 
\hline
 					 &  \expertlayerfont{Expert$^{(i)}\rightarrow$ CONV: 1$\times$64$\times$5}  \\ 
					 & \expertlayerfont{Expert$^{(i)}\rightarrow$ POOL: 3,2,Ave}  \\ 
\hline
\hline
\layerfont{FC: 64 } 	           & \expertlayerfont{Expert$^{(i)}\rightarrow$ FC: 64}  \\ 
\hline
\hline
\layerfont{\parbox[t]{3.5cm}{FC: $K$ (Generalist) \\ FC: $C$ (Base Model) }} 	           & \expertlayerfont{Expert$^{(i)}\rightarrow$ FC: $C/K$}  \\ 
\hline
\hline
\multicolumn{2}{|c|}{\bf Data preprocessing and augmentation} \\
\hline
\lrfont {
Input: $32\times32$ \\
Image mean subtraction \\
} & 
\lrfont {
Input: $32\times32$ \\
Image mean subtraction \\
}
\\
\hline
\multicolumn{2}{|c|}{\bf Learning Policy} \\
\hline
\lrfont {
\textit{Generalist:} \\ 
Learning rate: 0.001  (Fixed)  \\ 
0.001 : 8 Epochs  \\
Momentum: 0.9 \\
Weight decay: 0.004 \\ 
Weight initialization: Random \\
MiniBatch size: 100\\

\textit{Base Model:} \\ 
Learning rate: 0.001  (lowered twice)  \\ 
12 Epochs  \\
Momentum: 0.9 \\
Weight decay: 0.004 \\ 
Weight initialization: Random \\
MiniBatch size: 100 \\
 } 
 & 
 \lrfont {
 Learning rate: 0.001  till 0.00001 (lowered twice) \\
0.001 : 8 Epochs \\
0.0001 : 2 Epochs \\
0.00001 : 2 Epochs \\

Momentum: 0.9 \\
Weight decay: 0.004 \\
Weight initialization: Random \\
 MiniBatch size: 100 \\
 }

 \\
\hline 
\end{tabular}

\end{center}
\end{table}
\setlength{\tabcolsep}{1.4pt}

\setlength{\tabcolsep}{4pt}
\begin{table}[]
\begin{center}
\caption{\bf VGG11-C100 (trained on CIFAR100)}
\label{table:vgg_c100_config}

\begin{tabular}{ |p{5.5cm}|p{5.5cm}|  }
\hline

\cellcolor{blue!25} \bf{ Generalist and Base Model} &  \cellcolor{red!25} \bf{ Network of Experts (\nofe)} \\ 
\hline
\hline
 
\layerfont{CONV: 2$\times$64$\times$3 } & \layerfont{CONV: 2$\times$64$\times$3 }  \\ 
\layerfont{POOL: 2,2,Max}  & \layerfont{POOL: 2,2,Max}    \\
\hline
\hline 
\layerfont{CONV: 2$\times$128$\times$3 } & \layerfont{CONV: 2$\times$128$\times$3 }  \\ 
\layerfont{POOL: 2,2,Max}  & \layerfont{POOL: 2,2,Max}    \\
\hline
\hline
 \layerfont{CONV: 4$\times$256$\times$3 } & \layerfont{CONV: 4$\times$256$\times$3 }  \\ 
\layerfont{POOL: 2,2,Max}  & \layerfont{POOL: 2,2,Max}    \\
\hline 
\hline
				 & \expertlayerfont{Expert$^{(i)}\rightarrow$ CONV: 2$\times$256$\times$3  } \\ 
				&  \expertlayerfont{Expert$^{(i)}\rightarrow$ POOL: 2,2,Max}    \\
				 
\hline 
\hline
\layerfont{FC: $1024$}  & \expertlayerfont{Expert$^{(i)}\rightarrow$ FC: $512$ } \\ 
\hline 
\hline
\layerfont{FC: $1024$}    & \expertlayerfont{Expert$^{(i)}\rightarrow$ FC: $512$}  \\ 
\hline 
\hline
\layerfont{\parbox[t]{3.5cm}{FC: $K$ (Generalist) \\ FC: $C$ (Base Model) }}     & \expertlayerfont{Expert$^{(i)}\rightarrow$ FC: $C/K$}  \\ 
\hline
\hline
\multicolumn{2}{|c|}{\bf Data preprocessing and augmentation} \\
\hline
\lrfont {
4 zeros padding on each side, then $32\times32$ crops\\
Image mean subtraction \\
Image mirroring \\
} & 
\lrfont {
4 zeros padding on each side, then $32\times32$ crops\\
Image mean subtraction \\
Image mirroring \\
}
\\
\hline
\multicolumn{2}{|c|}{\bf Learning Policy} \\
\hline
\lrfont {
\textit{Generalist:} \\ 
Learning rate: 0.001  (Fixed)  \\ 
0.001 : 60 Epochs  \\
Momentum: 0.9 \\
Weight decay: 0.0005 \\ 
Weight initialization: Xavier \cite{GlorotEtAl:AISTATS2011} \\
MiniBatch size: 128 \\

\textit{Base Model:} \\ 
Learning rate: 0.001  (lowered twice)  \\ 
200 Epochs  \\
Momentum: 0.9 \\
Weight decay: 0.0005 \\ 
Weight initialization: Xavier \cite{GlorotEtAl:AISTATS2011} \\
MiniBatch size: 128 \\
} 
 & 
 \lrfont {
 Learning rate: 0.001  till 0.00001 (lowered twice) \\
0.001 : 120 Epochs \\
0.0001 : 10 Epochs \\
0.00001 : 10 Epochs \\

Momentum: 0.9 \\
Weight decay: 0.0005 \\
Weight initialization: Xavier \cite{GlorotEtAl:AISTATS2011} \\
 MiniBatch size: 128 \\}

 \\
\hline 

\end{tabular}

\end{center}
\end{table}
\setlength{\tabcolsep}{1.4pt}

\setlength{\tabcolsep}{4pt}
\begin{table}[]
\begin{center}
\caption{\bf NIN-C100 (trained on CIFAR100)}
\label{table:nin_c100_config}

\begin{tabular}{ |p{5.5cm}|p{5.5cm}|  }
\hline

\cellcolor{blue!25} \bf{ Generalist and Base Model} &  \cellcolor{red!25} \bf{ Network of Experts (\nofe)} \\ 
\hline
\hline
 
\layerfont{CONV: 1$\times$192$\times$5 } & \layerfont{CONV: 1$\times$192$\times$5 }  \\ 
\layerfont{CONV: 1$\times$160$\times$1 } & \layerfont{CONV: 1$\times$160$\times$1 }  \\ 
\layerfont{CONV: 1$\times$96$\times$1 } & \layerfont{CONV: 1$\times$96$\times$1 }  \\ 
\layerfont{POOL: 3,2,Max}  & \layerfont{POOL: 3,2,Max}    \\
\hline
\hline 
\layerfont{CONV: 1$\times$192$\times$5 } & \layerfont{CONV: 1$\times$192$\times$5 }  \\ 
\layerfont{CONV: 1$\times$192$\times$1 } & \layerfont{CONV: 1$\times$192$\times$1 }  \\ 
\layerfont{CONV: 1$\times$192$\times$1 } & \layerfont{CONV: 1$\times$192$\times$1 }  \\ 
\layerfont{POOL: 3,2,Max}  & \layerfont{POOL: 3,2,Max}    \\
\hline
\hline
\layerfont{CONV: 1$\times$192$\times$3 } & \layerfont{CONV: 1$\times$192$\times$3 }  \\ 
\layerfont{CONV: 1$\times$192$\times$1 } & \layerfont{CONV: 1$\times$192$\times$1 }  \\ 

\hline 
\hline
				 & \expertlayerfont{Expert$^{(i)}\rightarrow$ CONV: 1$\times$192$\times$3  } \\ 
				&  \expertlayerfont{Expert$^{(i)}\rightarrow$ CONV: 1$\times$192$\times$1 }    \\
\hline 
\hline
\layerfont{\parbox[t]{3.5cm}{CONV:1$\times$K$\times$1(Generalist)\\CONV:1$\times$C$\times$1(BaseModel)}}& \expertlayerfont{Expert$^{(i)}\rightarrow$ CONV: 1$\times$(C/K)$\times$1 }  \\ 
\hline
\hline
\layerfont{\parbox[t]{3.5cm}{POOL:8,1,AVE (Generalist)\\POOL:8,1,AVE (BaseModel)}}     & \expertlayerfont{Expert$^{(i)}\rightarrow$ POOL: 6,1,AVE }  \\ 
\hline
\hline
\multicolumn{2}{|c|}{\bf Data preprocessing and augmentation} \\
\hline
\lrfont {
Input crop: $26\times26$ \\
Image mean subtraction \\
Image mirroring \\
} & 
\lrfont {
Input crop: $26\times26$ \\
Image mean subtraction \\
Image mirroring \\
}
\\
\hline
\multicolumn{2}{|c|}{\bf Learning Policy} \\
\hline
\lrfont {
\textit{Generalist:} \\ 
Learning rate: 0.01  (Fixed)  \\ 
0.01 : 200 Epochs  \\
Momentum: 0.9 \\
Weight decay: 0.001 \\ 
Weight initialization:  Random \\
MiniBatch size: 100 \\

\textit{Base Model:} \\ 
Learning rate: 0.01  (lowered twice)  \\ 
0.01 : 220 Epochs  \\
0.001 : 10 Epochs  \\
0.0001 : 30 Epochs  \\
Momentum: 0.9 \\
Weight decay: 0.001 \\ 
Weight initialization: Random \\
MiniBatch size: 100 \\
} 
 & 
 \lrfont {
 Learning rate: 0.01  till 0.0001 (lowered twice) \\
0.01 : 98 Epochs \\
0.001 : 120 Epochs \\
0.0001 : 10 Epochs \\
Momentum: 0.9 \\
Weight decay: 0.001 \\ 
Weight initialization: Random \\
 MiniBatch size: 100 \\}

 \\
\hline 

\end{tabular}

\end{center}
\end{table}
\setlength{\tabcolsep}{1.4pt}


\setlength{\tabcolsep}{4pt}
\begin{table}[]
\begin{center}
\caption{\bf ResNet56-C100 (trained on CIFAR100)}
\label{table:res_c100_config}

\begin{tabular}{ |p{5.5cm}|p{5.8cm}|  }
\hline

\cellcolor{blue!25} \bf{ Generalist and Base Model} &  \cellcolor{red!25} \bf{ Network of Experts (\nofe)} \\ 
\hline
\hline
 
\layerfont{CONV: 1$\times$64$\times$3  } & \layerfont{CONV: 1$\times$64$\times$3 }  \\ 
\hline
\hline 
\layerfont{ \Bigg\{ \parbox[t]{2.5cm}{CONV: 1$\times$64$\times$3 \\ CONV: 1$\times$64$\times$3 \\}     \Bigg\}   $\times9$  } & 
\layerfont{ \Bigg\{ \parbox[t]{2.5cm}{CONV: 1$\times$64$\times$3 \\ CONV: 1$\times$64$\times$3 \\}     \Bigg\}   $\times9$  }  \\ 
\hline
\hline
\layerfont{ \Bigg\{ \parbox[t]{2.5cm}{CONV: 1$\times$128$\times$3 \\ CONV: 1$\times$128$\times$3 \\}     \Bigg\}   $\times9$  } & 
\layerfont{ \Bigg\{ \parbox[t]{2.5cm}{CONV: 1$\times$128$\times$3 \\ CONV: 1$\times$128$\times$3 \\}     \Bigg\}   $\times9$  }  \\ 
\hline
\hline
\layerfont{ \Bigg\{ \parbox[t]{2.5cm}{CONV: 1$\times$256$\times$3 \\ CONV: 1$\times$256$\times$3 \\}     \Bigg\}   $\times9$  } & 
\layerfont{ \Bigg\{ \parbox[t]{2.5cm}{CONV: 1$\times$256$\times$3 \\ CONV: 1$\times$256$\times$3 \\}     \Bigg\}   $\times9$  }  \\ 
\hline 
\hline
				 & \expertlayerfont{Expert$^{(i)}\rightarrow$ \Bigg\{ \parbox[t]{2.5cm}{CONV: 1$\times$256$\times$3 \\ CONV: 1$\times$256$\times$3 \\}     \Bigg\}   $\times1$   } \\ 
\layerfont{POOL: 7,1,Ave}  & \expertlayerfont{Expert$^{(i)}\rightarrow$ POOL: 7,1,Ave}    \\
\hline 
\hline
\layerfont{\parbox[t]{3.5cm}{FC: $K$ (Generalist) \\ FC: $C$ (Base Model) }}       & \expertlayerfont{Expert$^{(i)}\rightarrow$ FC: $C/K$}  \\

\hline
\hline
\multicolumn{2}{|c|}{\bf Data preprocessing and augmentation} \\
\hline
\lrfont {
Crop size: 28\\
Image mean subtraction \\
Image mirroring \\
} & 
\lrfont {
Crop size: 28\\
Image mean subtraction \\
Image mirroring \\
}
\\
\hline
\multicolumn{2}{|c|}{\bf Learning Policy} \\
\hline
\lrfont {
\textit{Generalist:} \\ 
Learning rate: 0.01  (Fixed)  \\ 
0.01 : 12 Epochs  \\
Momentum: 0.9 \\
Weight decay: 0.0001 \\ 
Weight initialization: MSRA \cite{HeEtAl:ICCV2015} \\
MiniBatch size: 128 \\ 

\textit{Base Model:} \\ 
Learning rate: 0.1  (lowered twice)  \\ 
60 Epochs  \\
Momentum: 0.9 \\
Weight decay: 0.0001 \\ 
Weight initialization: MSRA \cite{HeEtAl:ICCV2015} \\
MiniBatch size: 128 \\
} 
 & 
 \lrfont {
 Learning rate: 0.1  till 0.001 (lowered twice) \\
0.1 : 30 Epochs \\
0.01 : 15 Epochs \\
0.001 : 15 Epochs \\

Momentum: 0.9 \\
Weight decay: 0.0005 \\
Weight initialization: MSRA \cite{HeEtAl:ICCV2015} \\
 MiniBatch size: 128 \\}

 \\
\hline 

\end{tabular}

\end{center}
\end{table}
\setlength{\tabcolsep}{1.4pt}



\setlength{\tabcolsep}{4pt}
\begin{table}[]
\begin{center}
\caption{\bf AlexNet-Caffe (trained on ImageNet)}
\label{table:alex_imagenet_config}

\begin{tabular}{ |p{5.5cm}|p{5.5cm}|  }
\hline

\cellcolor{blue!25} \bf{ Generalist} &  \cellcolor{red!25} \bf{ Network of Experts (\nofe)} \\ 
\hline
\hline
 
\layerfont{CONV: 2$\times$96$\times$11 } & \layerfont{CONV: 2$\times$96$\times$11 }  \\ 
\layerfont{LRN } & \layerfont{LRN }  \\ 
\layerfont{POOL: 3,2,Max}  & \layerfont{POOL: 3,2,Max}    \\
\hline
\hline 
 
\layerfont{CONV: 2$\times$384$\times$3 } & \layerfont{CONV: 2$\times$384$\times$3 }  \\ 
\layerfont{CONV: 1$\times$256$\times$3 } & \layerfont{CONV: 1$\times$256$\times$3 }  \\ 

\layerfont{LRN } & \layerfont{LRN }  \\ 
\layerfont{POOL: 3,2,Max}  & \layerfont{POOL: 3,2,Max}    \\
\hline 
\hline
				 & \expertlayerfont{Expert$^{(i)}\rightarrow$ CONV: 1$\times$256$\times$3  } \\ 
				&  \expertlayerfont{Expert$^{(i)}\rightarrow$ POOL: 3,2,Max}    \\

\hline 
\hline
\layerfont{FC: $4096$}  & \expertlayerfont{Expert$^{(i)}\rightarrow$ FC: $1024$ } \\ 
\hline 
\hline
\layerfont{FC: $4096$}    & \expertlayerfont{Expert$^{(i)}\rightarrow$ FC: $1024$}  \\ 
\hline 
\hline
\layerfont{\parbox[t]{3.5cm}{FC: $K$ (Generalist) \\ FC: $C$ (Base Model) }}       & \expertlayerfont{Expert$^{(i)}\rightarrow$ FC: $C/K$}  \\

\hline
\hline
\multicolumn{2}{|c|}{\bf Data preprocessing and augmentation} \\
\hline
\lrfont {
Crop size: 227\\
Image mean subtraction \\
Image mirroring \\
} & 
\lrfont {
Crop size: 227\\
Image mean subtraction \\
Image mirroring \\

}
\\
\hline
\multicolumn{2}{|c|}{\bf Learning Policy} \\
\hline
\lrfont {
\textit{Generalist:} \\ 
Learning rate: 0.01  (Fixed)  \\ 
0.01: 20 Epochs  \\
Momentum: 0.9 \\
Weight decay: 0.0005 \\ 
Weight initialization: Random \\
MiniBatch size: 256 \\

\textit{Base Model:} \\ 
Learning rate: 0.01  (lowered twice)  \\ 
80 Epochs  \\
Momentum: 0.9 \\
Weight decay: 0.0005 \\ 
Weight initialization: Random \\
MiniBatch size: 256 \\

 } 
 & 
 \lrfont {
 Learning rate: 0.01  till 0.0001 (lowered twice) \\
0.001 : 20 Epochs \\
0.001 : 20 Epochs \\
0.0001 : 20 Epochs \\

Momentum: 0.9 \\
Weight decay: 0.0005 \\
Weight initialization: Random \\
 MiniBatch size: 256 \\}

 \\
\hline 

\end{tabular}

\end{center}
\end{table}
\setlength{\tabcolsep}{1.4pt}

\clearpage
\subsection{Software Implementation}
\label{software_section}

Our software implementation is based on the Caffe library~\cite{caffe}. In order to implement our Network of Experts, we have made changes to the Caffe library, including new development of layers, solvers, and tools. Software implementing our method and several pretrained \nofe models are available at \url{http://vlg.cs.dartmouth.edu/projects/nofe/}.

\bibliographystyle{splncs03}
\bibliography{egbib}

\end{document}